\documentclass{article}
\usepackage{iclr2023_conference,times}

\usepackage{amsmath,amsfonts,bm}

\def\eqref#1{equation~\ref{#1}}

\def\1{\bm{1}}

\DeclareMathAlphabet{\mathsfit}{\encodingdefault}{\sfdefault}{m}{sl}
\SetMathAlphabet{\mathsfit}{bold}{\encodingdefault}{\sfdefault}{bx}{n}

\newcommand{\R}{\mathbb{R}}

\usepackage{hyperref}
\usepackage{url}
\usepackage{graphicx}
\usepackage{amsmath}
\usepackage{multirow}
\usepackage{subcaption}

\usepackage{diagbox}

\title{AnomalyBERT: Self-Supervised Transformer for Time Series Anomaly Detection using Data Degradation Scheme}
\author{Yungi Jeong, Eunseok Yang, Jung Hyun Ryu, Imseong Park, Myungjoo Kang\thanks{Corresponding author.} \\
Numerical Computing \& Image Analysis Lab\\
Seoul National University, Seoul, Republic of Korea \\
\texttt{\{jyg9628,mayth24,jhryu30,parkis,mkang\}@snu.ac.kr}
}

\def\genbox#1#2#3#4#5#6{
    \leavevmode\raise#4bp\hbox to#5bp{\vrule height#5bp depth0bp width0bp
    \pdfliteral{q .5 w \csname #2COLOR\endcsname\space RG
                       \csname #3PDF\endcsname{#5}{#6} S Q
             \ifx1#1 q \csname #2COLOR\endcsname\space rg 
                       \csname #3PDF\endcsname{#5}{#6} f Q\fi}\hss}}

\def\circbox    #1#2{\genbox{#1}{#2}  {circ}     {0}   {5}    {2.5}}

\definecolor{lightred}{cmyk}{0, 0.7808, 0.4429, 0.1412}

\iclrfinalcopy
\begin{document}

\maketitle

\begin{abstract}
Mechanical defects in real situations affect observation values and cause abnormalities in multivariate time series, such as sensor values or network data.
To perceive abnormalities in such data, it is crucial to understand the temporal context and interrelation between variables simultaneously.
The anomaly detection task for time series, especially for unlabeled data, has been a challenging problem, and we address it by applying a suitable {\it data degradation scheme} to self-supervised model training.
We define four types of synthetic outliers and propose the degradation scheme in which a portion of input data is replaced with one of the synthetic outliers.
Inspired by the self-attention mechanism, we design a Transformer-based architecture to recognize the temporal context and detect unnatural sequences with high efficiency.
Our model converts multivariate data points into temporal representations with relative position bias and yields anomaly scores from these representations.
Our method, {\it AnomalyBERT}, shows a great capability of detecting anomalies contained in complex time series and surpasses previous state-of-the-art methods on five real-world benchmarks.
Our code is available at \url{https://github.com/Jhryu30/AnomalyBERT}.
\end{abstract}

\begin{figure}[h]
\begin{center}
\includegraphics[width=0.98\textwidth]{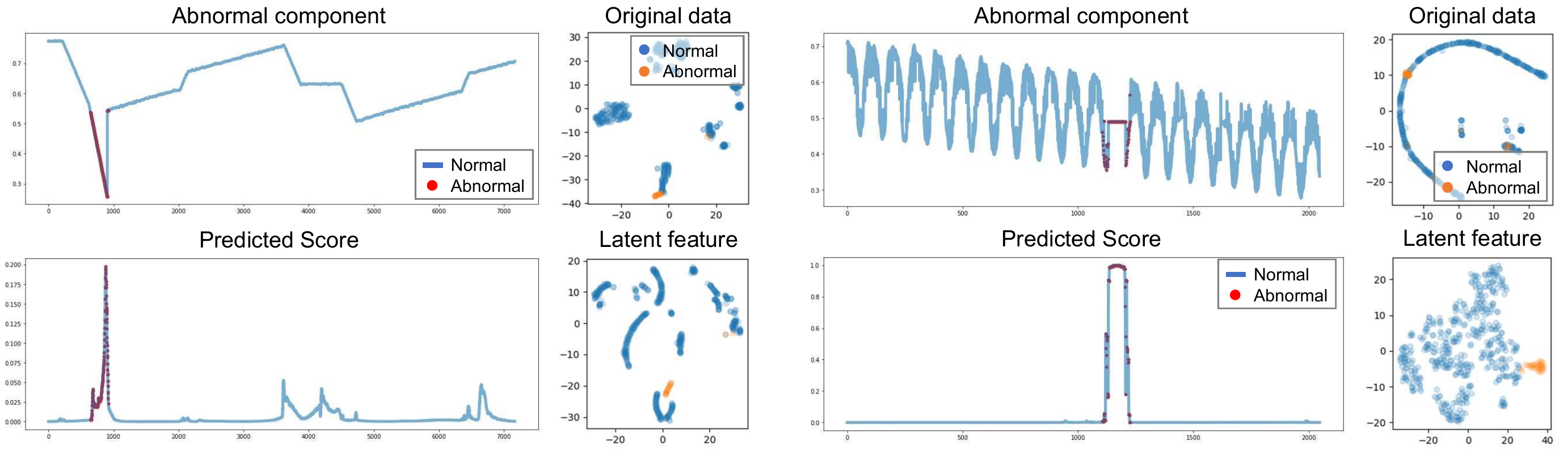}
\end{center}
\setlength{\abovecaptionskip}{5pt plus 3pt minus 2pt}
\caption{Examples of anomaly detection for abnormal time series in SWaT ({\it left}) and SMAP ({\it right}) datasets. Our method, AnomalyBERT, predicts anomaly scores that quantify abnormalities of data points in time series. We also visualize the original data and the final latent features in our model using t-SNE and show that our method separates abnormal points effectively.}
\label{fig:introduction_examples}
\end{figure}

\section{Introduction}

In many industrial environments, time series data is mainly dealt with to monitor machines, IT devices, spacecrafts, or engines.
Anomaly detection is one of the essential tasks for time series analysis, which can find defects in machines and prevent potential harm.
Recently, many deep learning-based approaches have been applied to this work.
Several studies design recurrent neural network (RNN) models \citep{hundman2018detecting, park2018multimodal, su2019robust} to treat multivariate data in the temporal order.
Some researchers try to adopt a graph or Transformer architecture \citep{vaswani2017attention} to focus on relationships between variables and data points \citep{deng2021graph, xu2021anomaly}.
These approaches take account of the temporal characteristics of data and successfully adapt deep neural networks to the field of time series analysis.

Most datasets do not provide ground truth labels for the training set in this area.
In other words, it is unknown whether a point is anomalous or not in the training set.
Therefore, previous studies have developed unsupervised learning methods for anomaly detection.
Some of them insert an autoencoder or an adversarial network \citep{goodfellow2014generative} into their model to understand data distribution efficiently.
However, it is still a hard problem to detect anomalies from the temporal context without supervision.
Moreover, abnormalities may display unexpected behavior and be related to multiple variables, which makes the detection task more difficult in real situations.

In this paper, we design a Transformer-based architecture and propose {\it AnomalyBERT}, a self-supervised method for time series anomaly detection.
Inspired by BERT \citep{devlin2018bert} in the natural language processing (NLP) field, we modify the masked language modeling (MLM) by replacing a random portion of input data and training a model to find the degraded part.
This {\it data degradation scheme} helps detect varied unnatural sequences in real-world time series, as shown in Figure \ref{fig:introduction_examples}.
Furthermore, we apply 1D relative position bias \citep{raffel2019exploring} to self-attention modules to insert appropriate temporal information into data.
AnomalyBERT outperforms previous detection methods by achieving the highest F1-scores on five real-world benchmark datasets.
We demonstrate that our data degradation scheme enables the Transformer-based model to understand the temporal context, and our method has strong capability in detecting real-world anomalies.


\section{Related Works}

{\bf Time Series Anomaly Detection.}
Anomaly detection problems have been handled using various statistical and machine learning-based (ML-based) methods.
A classical method, Local Outlier Factor (LOF) \citep{breunig2000lof}, is proposed for density-based outlier detection.
Following the LOF, several statistical methods \citep{tang2002enhancing, kriegel2009loop} and ML-based methods \citep{tax2004support, ruff2018deep, liu2008isolation} have been proposed.
For example, DAGMM \citep{zong2018deep} combines Gaussian Mixture Model (GMM) with a deep neural network.  
Meanwhile, deep learning-based approaches have appeared in this area with the advances in neural networks.
They commonly adopt RNNs to deal with complex time series, such as multiple sensor values from IoT data.
\citet{park2018multimodal} propose LSTM-VAE, a variational autoencoder (VAE) model whose feed-forward networks are replaced with Long Short-Term Memory (LSTM). 
\citet{su2019robust} also propose a VAE model, but they use a gated recurrent unit (GRU) to extract latent features.
Recently, there have been attempts to employ advanced models. 
\citet{deng2021graph} present a graph neural network (GNN) that detects deviations from the relationships between variables, and \citet{xu2021anomaly} propose a Transformer-based model and define the association discrepancy for detection criterion.

\smallskip
{\bf Transformer and Its Variants.}
Transformer \citep{vaswani2017attention} is first introduced in the field of NLP and has achieved big success.
It employs an attention mechanism in the multi-head structures and builds a pair of an encoder and a decoder with the attention layers.
The concepts of Transformer have been applied to various NLP tasks.
For example, BERT \citep{devlin2018bert} uses a Transformer encoder that is pre-trained on large-scale unlabeled datasets.
BERT implements MLM in the pre-training phase to understand the context of the sentences.
There also have been successful methods for effective generalization.
T5 \citep{raffel2019exploring} is a text-to-text method using both an encoder and a decoder.
SpanBERT \citep{joshi2020spanbert} introduces a span masking instead of the MLM in BERT, and XLNet \citep{yang2019xlnet} combines autoencoding and autoregressive modeling.
BART \citep{lewis2019bart} uses several noising schemes including the MLM in BERT.
The Transformer architecture also has been applied to computer vision tasks recently.
Vision Transformer (ViT) \citep{dosovitskiy2020image} employs a Transformer encoder without CNN architecture and achieves outstanding results in classification tasks.


\begin{figure}[t]
\begin{center}
\begin{subfigure}{0.79\textwidth}
\centering
\includegraphics[width=\linewidth]{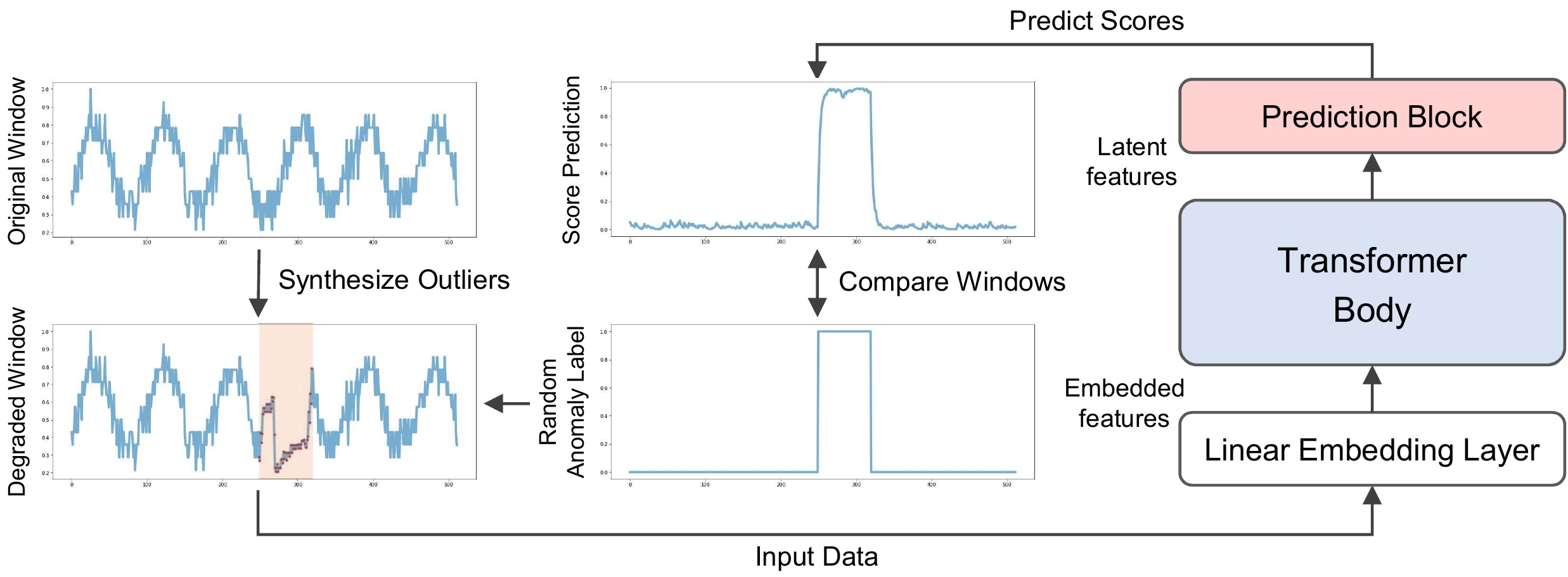}
\end{subfigure}
\begin{subfigure}{0.185\textwidth}
\centering
\includegraphics[width=\linewidth]{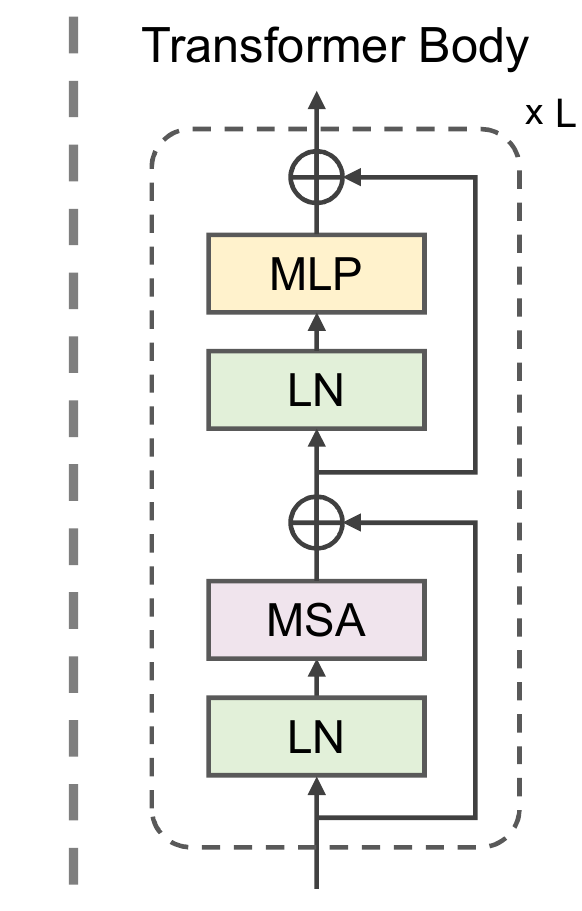}
\end{subfigure}
\end{center}
\setlength{\abovecaptionskip}{5pt plus 3pt minus 2pt}
\caption{An overview of our framework. We design a Transformer-based model and a self-supervised training strategy. In the training stage, a portion of an input window is randomly replaced and the model is directed to classify the degraded part. The main Transformer body is composed of Transformer layers with 1D relative position bias.}
\label{fig:overview}
\end{figure}

\section{Method}


\subsection{Overall Architecture}

An overview of our framework is illustrated in Figure \ref{fig:overview}.
Our model is composed of three parts; a linear embedding layer, a Transformer body, and a prediction block.
A window of multivariate time series $X = x_{t_0:t_1} \in \R^{N \times D}$ is fed into the model as an input.
The linear embedding layer first projects each data patch $x_{t:t+p}$ (a patch consists of several neighboring points) in a window $X$ to an embedded feature $f_i$.
Then the Transformer body takes all embedded features $\{f_i\}_{1 \le i \le M}$ from $X$ and yields latent features $\{h_i\}_{1 \le i \le M}$.
These latent features share information among themselves and reflect the temporal context in the window.
The prediction block finally outputs anomaly scores of data points $a_{t_0:t_1} \in [0, 1]^{N}$ of the window.
A data point $x_t$ is regarded as more anomalous as the score $a_t$ is closer to 1.


We adopt the Transformer encoder, in which each layer contains a multi-head self-attention (MSA) module and an MLP block, as the main body.
A LayerNorm (LN) layer is placed before each module, and GELU activation is used for activation layers.
Unlike the original Transformer or ViT, we do not use sinusoidal positional encodings \citep{vaswani2017attention} or absolute position embeddings \citep{dosovitskiy2020image} to inject positional information.
We instead add 1D relative position bias \citep{raffel2019exploring, liu2021swin} to each attention matrix to consider the relative positions between features in a window. 
A self-attention in each head with the relative bias is computed as:
\begin{equation}
  \textrm{Attention}(Q, K, V) = \textrm{SoftMax}\left(\frac{QK^T}{\sqrt{d}} + B \right) V,
  \label{eq:attention}
\end{equation}
\noindent where $Q$, $K$, and $V$ are query, key, and value of input features, respectively, and $d$ is the dimension of features in an attention head.
$B = [b_{i,j}] \in \mathbb{R}^{M \times M}$ is a relative position bias and an element $b_{i,j} = \hat{b}_{j-i}$ is brought from a learnable bias table $\hat{B} = \{\hat{b}_n\}_{-M+1 \le n \le M-1}$.
We apply a different position bias to each MSA module as in \citet{liu2021swin}.

\subsection{Synthetic Outliers and Data Degradation}

\begin{figure}[t]
\begin{center}
\includegraphics[width=0.98\textwidth]{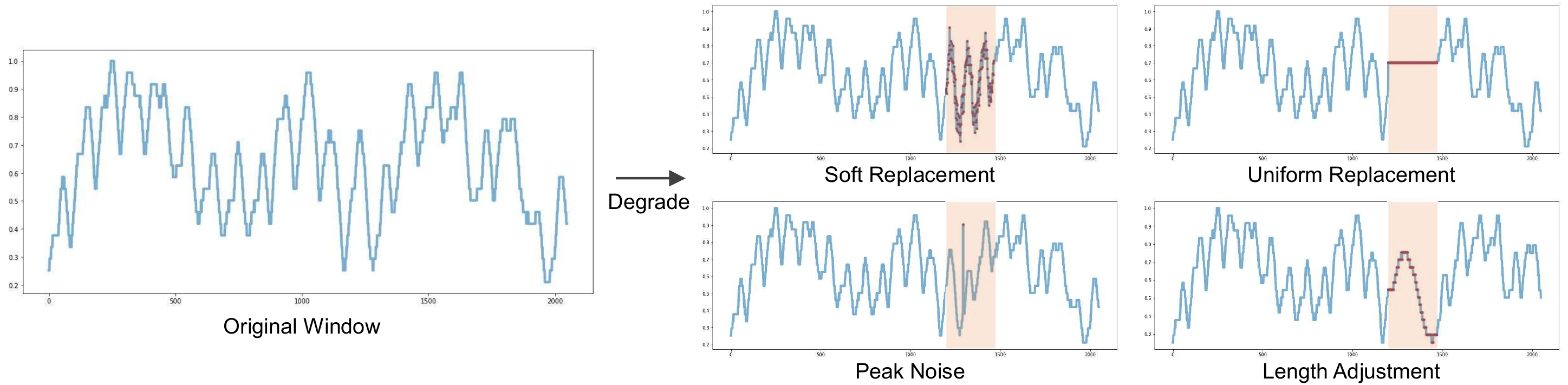}
\end{center}
\setlength{\abovecaptionskip}{5pt plus 3pt minus 2pt}
\caption{Types of synthetic outliers and degradation examples. We define four types of synthetic outliers, soft replacement, uniform replacement, peak noise, and length adjustment, which are added to input windows for the model training.}
\label{fig:synthetic_outlier}
\end{figure}

As we use unlabeled training data, we create degraded inputs by replacing a portion of a window with an outlier in the training phase (Figure \ref{fig:synthetic_outlier}).
Similar to the span masking in SpanBERT \citep{joshi2020spanbert}, we randomly select an interval $[t^{'}_0, t^{'}_1] \subset [t_0, t_1]$ in a window $X = x_{t_0:t_1}$. 
The selected sequence $X^{'} = x_{t^{'}_0:t^{'}_1}$ is replaced with one of the synthetic outliers below.
\vspace*{-2mm}
\begin{itemize}
    \setlength\itemsep{-0.2mm}
    \item A weighted sequence with the outside of the window (Soft replacement).
    \item A constant sequence (Uniform replacement).
    \item A lengthened or shortened sequence (Length adjustment).
    \item A single peak value (Peak noise).
\end{itemize}
The {\it soft replacement} denotes the replaced sequence fetched from the outside of the window.
Technically, it represents the replacement with a weighted sum of the original interval and an external interval.
The {\it uniform replacement} is the replacement with a constant sequence, and the {\it length adjustment} denotes a lengthened or shortened sequence.
Lastly, the {\it peak noise} is the addition of a single peak value.
Unlike the existing method \citep{lai2021revisiting}, our data degradation scheme can be processed without prior knowledge of a given time series.

\subsection{Training}

We apply the binary cross entropy loss to our objective.
For an input window $X = x_{t_0:t_1}$ with a degraded interval $[t^{'}_0, t^{'}_1]$ and predicted anomaly scores $a_{t_0:t_1}$, the objective function is defined as:
\begin{equation}
  L = - \frac{1}{N} \sum_{t=t_0}^{t_1} \1_\mathrm{[t^{'}_0, t^{'}_1]}(t) \cdot \log a_t + \left(1 - \1_\mathrm{[t^{'}_0, t^{'}_1]}(t) \right) \cdot \log (1 - a_t),
  \label{eq:objective}
\end{equation}
\noindent where $N = t_1 - t_0 + 1$ is the window size.
The function $\1_\mathrm{[t^{'}_0, t^{'}_1]}$ plays a role of artificial labels in this equation.
Compared to the MLM in the field of NLP \citep{devlin2018bert, yang2019xlnet}, our model is directed to classify the entire data points in a window into normal/abnormal points at once.
At every training step, a synthetic outlier of random type, length, and values is added to an original window under the data degradation scheme.
A mini-batch of degraded windows is fed into the model, and the model is trained to detect the degraded parts.
The implementation details of the training procedure are described in Appendix \ref{sec:appendix_settings}.


\section{Experiments}


\subsection{Datasets}
\label{sec:datasets}
We create a simple sine wave dataset that consists of a normal sequence and five abnormal sequences categorized by \citet{lai2021revisiting}, and conduct a preliminary experiment on this dataset.
We then produce the experimental results on five widely-used benchmark datasets, SWaT, WADI, SMAP, MSL, and SMD \citep{goh2017dataset, ahmed2017wadi, hundman2018detecting, su2019robust}.
These datasets are collected from multiple sensors in server machines, spacecrafts, or water treatment/distribution systems.
Each dataset consists of an unlabeled training set and a labeled test set.
The information of datasets is summarized in Table \ref{tab:data_info} and described in Appendix \ref{sec:appendix_datasets} in detail.

\subsection{Score Prediction and Evaluation Metrics}
In the evaluation stage, the trained model takes windows in the test set and predicts anomaly scores of the data points.
We use the sliding window strategy \citep{shen2020timeseries} and average the scores on the overlapped intervals.
After the score prediction, we categorize data points whose anomaly scores exceed a threshold as anomalies.
We mainly use F1-score ($\mathrm{F1}$) over the ground-truth labels and anomaly predictions to evaluate the performance.
We count the number of true positives (TP), false positives (FP), and false negatives (FN), and compute $\mathrm{F1}$ as $2\text{TP} \,/\, (2\text{TP} + \text{FP} + \text{FN})$. 
In practice, many approaches process prediction results using the point adjustment \citep{xu2018unsupervised}, in which the entire points in an abnormal segment are regarded as anomalous if at least one point is detected as an anomaly.
This process, however, has been shown to overestimate the detection performance because it may increase TP but decrease FN dramatically \citep{kim2022towards}.
Following the protocol in \citet{kim2022towards}, thresholds that yield the best $\mathrm{F1}$ and F1-score after the point adjustment ($\mathrm{F1}_\mathrm{PA}$) are obtained, and the best evaluation values are used for comparisons.

\begin{table}[t]
\renewcommand{\arraystretch}{1.1}
\caption{Examination of relationships between the typical types of outliers in \citet{lai2021revisiting} and our proposed synthetic outliers. 
\circbox0{cblue} represents the covering of typical outlier types. It is considered to be covered if both $\mathrm{F1}$ and AUROC score are more than 0.9.}
\label{tab:Motivation_results}
\vspace*{-2mm}
\begin{center}
\begin{tabular}{l|ccccc}
\multirow{2}{*}{\diagbox{Proposed}{Typical}}
& \multicolumn{2}{c}{\bf Point} & \multicolumn{3}{c}{\bf Pattern} \\
                    &       Global        &    Contextual   &     Shapelet     &     Seasonal      &     Trend          \\ \hline
Soft replacement    & \circbox0{cblue}   & \circbox0{cblue} & \circbox0{cblue} & \circbox0{cblue}  &  \circbox0{cblue}  \\
Uniform replacement &                    & \circbox0{cblue} & \circbox0{cblue} &                   &                     \\
Length adjustment   &                    &                  &                  &  \circbox0{cblue} &                     \\
Peak noise          &  \circbox0{cblue}  & \circbox0{cblue} &                  &                   &      
\end{tabular}
\end{center}
\end{table}

\begin{figure}[t]
\begin{center}
\includegraphics[width=0.98\textwidth]{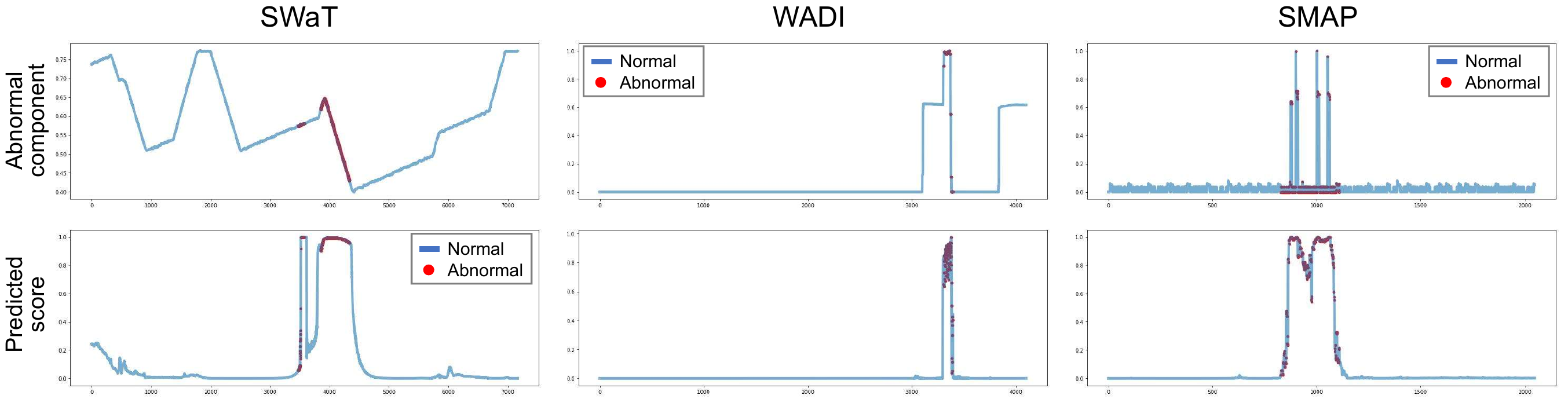}
\end{center}
\setlength{\abovecaptionskip}{5pt plus 3pt minus 2pt}
\caption{Qualitative results of AnomalyBERT on abnormal sequences from SWaT, WADI, and SMAP datasets. We visualize an abnormal component where the outlier occurs and anomaly scores for each sequence. Without any post-processing, our method finds out diverse types of anomalies.}
\label{fig:qualitative_results}
\end{figure}

\subsection{Motivation}
\label{sec:motivation}
Previous work \citep{lai2021revisiting} categorizes sequential outliers into five specific behavior-driven taxonomy; global, contextual, shapelet, seasonal, and trend outliers.
Synthesizing such patterns in time series data is a difficult task because it strictly requires knowledge of patterns that appear in existing data in advance.
For example, it is necessary to select a shape to be added to synthesize a shapelet outlier, and the data trend is required to synthesize a trend outlier.
However, our synthesis technique covers these five outliers in an easier way because it does not require analysis of the original data. 
We now refer to these five outlier types as the {\it typical outlier types}.

In Table \ref{tab:Motivation_results}, we examine the relationships between the typical outlier types and our synthetic outliers using a simplified version of our model and the sine wave dataset.
We measure $\mathrm{F1}$ and area under ROC curve (AUROC) for comparisons of the model performances,
and consider that a synthetic outlier {\it covers} a typical outlier type (\circbox0{cblue} mark) if a model trained with the synthetic one achieves both $\mathrm{F1}$ and AUROC over 0.9 on the sine wave including the corresponding typical outlier.  
As shown in Table \ref{tab:Motivation_results}, the soft replacement covers all typical types of outliers, and the uniform replacement and peak noise partially cover them.
Meanwhile, the length adjustment only covers the seasonal outlier.
From the simple experiment, we draw inspiration and get ideas for imitating anomalous behavior using synthetic outliers.


\subsection{Main Results}
\label{sec:main_results}

We report the results of AnomalyBERT on the five real-world datasets introduced in Section \ref{sec:datasets}, and compare them to those of the previous works.
The reproduced evaluation scores are brought from \citet{kim2022towards}.
In Table \ref{tab:F1_main_results}, we show that AnomalyBERT outperforms the previous methods on all datasets with $\mathrm{F1}$.
Our method particularly surpasses the others on MSL and SMAP, which may contain unlabeled outliers in the training set.
Despite the difficulty in training networks, our method detects anomalies well in this kind of data.
AnomalyBERT also performs well with $\mathrm{F1}_\mathrm{PA}$, although $\mathrm{F1}_\mathrm{PA}$ tends to distort the model performance.
Our qualitative results are visualized in Figure \ref{fig:qualitative_results}.

\begin{table}[t]
\setlength{\tabcolsep}{5pt}
\renewcommand{\arraystretch}{1.1}
\caption{F1-scores for various anomaly detection methods and AnomalyBERT on five benchmark datasets. We report the standard $\mathrm{F1}$ and F1-scores after point adjustment ($\mathrm{F1}_\mathrm{PA}$) following the protocol in \citet{kim2022towards}. Our method outperforms all existing methods with $\mathrm{F1}$.}
\label{tab:F1_main_results}
\vspace*{-2mm}
\begin{center}
{\small
\begin{tabular}{l|cccccccccc}
& \multicolumn{2}{c}{\bf SWaT} & \multicolumn{2}{c}{\bf WADI} & \multicolumn{2}{c}{\bf MSL} & \multicolumn{2}{c}{\bf SMAP} & \multicolumn{2}{c}{\bf SMD} \\
& $\mathrm{F1}$ & $\mathrm{F1}_\mathrm{PA}$ & $\mathrm{F1}$ & $\mathrm{F1}_\mathrm{PA}$ & $\mathrm{F1}$ & $\mathrm{F1}_\mathrm{PA}$ & $\mathrm{F1}$ & $\mathrm{F1}_\mathrm{PA}$ & $\mathrm{F1}$ & $\mathrm{F1}_\mathrm{PA}$ \\
\hline
DAGMM \citeyearpar{zong2018deep} & 0.550 & 0.853 & 0.121 & 0.209 & 0.199 & 0.701 & \underline{0.333} & 0.712 & 0.238 & 0.723 \\
LSTM-VAE \citeyearpar{park2018multimodal} & 0.775 & 0.805 & 0.227 & 0.380 & 0.212 & 0.678 & 0.235 & 0.756 & 0.435 & 0.808 \\
OmniAnomaly \citeyearpar{su2019robust} & 0.782 & 0.866 & 0.223 & 0.417 & 0.207 & 0.899 & 0.227 & 0.805 & 0.474 & {\bf 0.944} \\
MSCRED \citeyearpar{zhang2019deep} & 0.662 & 0.868 & 0.087 & 0.346 & 0.199 & 0.775 & 0.232 & {\bf 0.945} & 0.097 & 0.389 \\
THOC \citeyearpar{shen2020timeseries} & 0.612 & 0.880 & 0.130 & 0.506 & 0.190 & 0.891 & 0.240 & 0.781 & 0.168 & 0.541 \\
USAD \citeyearpar{audibert2020usad} & 0.791 & 0.846 & 0.232 & 0.429 & 0.211 & {\bf 0.927} & 0.228 & 0.818 & 0.426 & \underline{0.938} \\
GDN \citeyearpar{deng2021graph} & \underline{0.808} & {\bf 0.935} & \underline{0.569} & {\bf 0.855} & \underline{0.217} & \underline{0.903} & 0.252 & 0.708 & \underline{0.529} & 0.716 \\
AnomalyBERT (Ours) & {\bf 0.854} & \underline{0.925} & {\bf 0.580} & \underline{0.798} & {\bf 0.302} & 0.585 & {\bf 0.457} & \underline{0.914} & {\bf 0.535} & 0.830
\end{tabular}
}
\end{center}
\end{table}

\subsection{Impact of Synthetic Outliers}

\begin{table}[t]
\caption{Results of ablation studies on combinations of the synthetic outlier types. We show the impacts of synthetic outliers by comparing F1-scores on various experimental conditions.}
\label{tab:ablation_study}
\vspace*{-1mm}
{\small
\parbox{0.6\linewidth}{
\begin{subtable}{\linewidth}
\setlength{\tabcolsep}{5pt}
\renewcommand{\arraystretch}{1.1}
\caption{Experimental results on the soft replacement, uniform replacement, and peak noise on {\bf WADI} dataset.}
\label{tab:ablation_1}
\vspace{-1mm}
\centering
\begin{tabular}{ccc|cc}
\vspace{-1mm}
Soft & Uniform & \multirow{2}{*}{Peak noise} & \multirow{2}{*}{\hspace{1pt} $\mathrm{F1}$} & \multirow{2}{*}{\hspace{1pt} $\mathrm{F1}_\mathrm{PA}$} \\
replacement & replacement && \\
\hline
\circbox0{cblue} & \circbox0{cblue} & \circbox0{cblue} & \hspace{1pt} {\bf 0.580}     & \underline{0.798} \\
\circbox0{cblue} & \circbox0{cblue} & \textcolor{lightred}{$\times$} & \hspace{1pt} 0.504           & 0.756          \\
\circbox0{cblue} & \textcolor{lightred}{$\times$} & \circbox0{cblue} & \hspace{1pt} \underline{0.556} & 0.770          \\
\textcolor{lightred}{$\times$} & \circbox0{cblue} & \circbox0{cblue} & \hspace{1pt} 0.402           & 0.757          \\
\circbox0{cblue} & \textcolor{lightred}{$\times$} & \textcolor{lightred}{$\times$} & \hspace{1pt} 0.478           & 0.743          \\
\textcolor{lightred}{$\times$} & \circbox0{cblue} & \textcolor{lightred}{$\times$} & \hspace{1pt} 0.403           & 0.706          \\
\textcolor{lightred}{$\times$} & \textcolor{lightred}{$\times$} & \circbox0{cblue} & \hspace{1pt} 0.330      & {\bf 0.888}    
\end{tabular}
\end{subtable}
}
\hfill
\parbox{0.38\linewidth}{
\begin{subtable}{\linewidth}
\setlength{\tabcolsep}{5pt}
\renewcommand{\arraystretch}{1.1}
\centering
\caption{Experimental results on the length adjustment on {\bf SWaT} dataset.}
\label{tab:ablation_2_1}
\vspace{-1mm}
\begin{tabular}{c|cc}
Length adjustment & \hspace{1pt} $\mathrm{F1}$ & \hspace{1pt} $\mathrm{F1}_\mathrm{PA}$ \\
\hline
\circbox0{cblue} & \hspace{1pt} {\bf 0.854} & {\bf 0.925} \\
\textcolor{lightred}{$\times$} & \hspace{1pt} 0.837 & 0.914
\end{tabular}

\vspace{0.2cm}

\caption{Experimental results on the length adjustment on {\bf WADI} dataset.}
\label{tab:ablation_2_2}
\vspace{-1mm}
\begin{tabular}{c|cc}
Length adjustment & \hspace{1pt} $\mathrm{F1}$ & \hspace{1pt} $\mathrm{F1}_\mathrm{PA}$ \\
\hline
\circbox0{cblue} & \hspace{1pt} 0.433 & 0.642 \\
\textcolor{lightred}{$\times$} & \hspace{1pt} {\bf 0.580} & {\bf 0.798}
\end{tabular}
\end{subtable}
}}
\end{table}

We conduct ablation studies on various synthetic outliers in the training phase and show their impacts on the model performance.
In Table \ref{tab:ablation_1}, we first report the model performance affected by the three types of outliers, soft replacement, uniform replacement, and peak noise on WADI dataset.
We set a baseline by mixing all three outlier types and set experimental conditions by excluding outliers from the baseline one by one.
The sum of all probability of synthesizing outliers is fixed at 80\%. 
As shown in Table \ref{tab:ablation_1}, using all three outliers yields the best $\mathrm{F1}$, and using the soft replacement and peak noise yields the next.
The absence of soft replacement obviously reduces the capability of the model.
Also noteworthy, mixing other outlier types with soft replacement enhances the performance compared to using it only.
This indicates that the uniform replacement and peak noise complement the soft replacement, though each of them does not perfectly cover the typical outliers in Table \ref{tab:Motivation_results}.

On the other hand, the length adjustment has different influences depending on the datasets.
In Table \ref{tab:ablation_2_1} and \ref{tab:ablation_2_2}, we report the results of ablation studies on the length adjustment on SWaT and WADI datasets.
Using the length adjustment in the training stage enhances the model performance on SWaT dataset, but it degrades that on WADI dataset.
Because the length adjustment specializes in detecting abnormal frequencies as shown in Table \ref{tab:Motivation_results}, it may confuse the model if data contains various frequencies (SMAP, MSL) or low-frequencies (WADI).


\section{Conclusion}

This paper presents AnomalyBERT, a novel method for time series anomaly detection that uses a data degradation scheme to train a Transformer-based model in a self-supervised manner.
We design an appropriate Transformer architecture with 1D relative position embeddings for temporal data and propose four types of synthetic outliers that can cover all typical types of anomalies.
Exploiting the synthetic outliers in the training phase, our proposed model can learn to distinguish anomalous behavior.
We finally show that our method outperforms previous works and has a strong capability in detecting real-world anomalies in complex time series.
Our data degradation scheme has the potential to improve the model performance by revising degradation algorithms to mimic real-world anomalies naturally or mixing proper types of outliers according to data characteristics.
Therefore, future studies could be demonstrated on detailed analysis of outlier synthesis processes.

\subsubsection*{Acknowledgments}
This work was partly supported by Institute of Information \& Communications Technology
Planning \& Evaluation (IITP) grant funded by the Korea government (MSIT) [NO.2021-0-01343, Artificial Intelligence Graduate School Program (Seoul National University)], the NRF grant [2021R1A2C3010887], and the ICT R\&D program of MSIT/IITP [1711117093, 2021-0-00077] and MOTIE [P0014715].

\bibliography{iclr2023_conference}

\begin{thebibliography}{30}
\providecommand{\natexlab}[1]{#1}
\providecommand{\url}[1]{\texttt{#1}}
\expandafter\ifx\csname urlstyle\endcsname\relax
  \providecommand{\doi}[1]{doi: #1}\else
  \providecommand{\doi}{doi: \begingroup \urlstyle{rm}\Url}\fi

\bibitem[Ahmed et~al.(2017)Ahmed, Palleti, and Mathur]{ahmed2017wadi}
Chuadhry~Mujeeb Ahmed, Venkata~Reddy Palleti, and Aditya~P Mathur.
\newblock Wadi: a water distribution testbed for research in the design of
  secure cyber physical systems.
\newblock In \emph{Proceedings of the 3rd international workshop on
  cyber-physical systems for smart water networks}, pp.\  25--28, 2017.

\bibitem[Audibert et~al.(2020)Audibert, Michiardi, Guyard, Marti, and
  Zuluaga]{audibert2020usad}
Julien Audibert, Pietro Michiardi, Fr{\'e}d{\'e}ric Guyard, S{\'e}bastien
  Marti, and Maria~A Zuluaga.
\newblock Usad: Unsupervised anomaly detection on multivariate time series.
\newblock In \emph{Proceedings of the 26th ACM SIGKDD International Conference
  on Knowledge Discovery \& Data Mining}, pp.\  3395--3404, 2020.

\bibitem[Breunig et~al.(2000)Breunig, Kriegel, Ng, and Sander]{breunig2000lof}
Markus~M Breunig, Hans-Peter Kriegel, Raymond~T Ng, and J{\"o}rg Sander.
\newblock Lof: identifying density-based local outliers.
\newblock In \emph{Proceedings of the 2000 ACM SIGMOD international conference
  on Management of data}, pp.\  93--104, 2000.

\bibitem[Deng \& Hooi(2021)Deng and Hooi]{deng2021graph}
Ailin Deng and Bryan Hooi.
\newblock Graph neural network-based anomaly detection in multivariate time
  series.
\newblock In \emph{Proceedings of the AAAI Conference on Artificial
  Intelligence}, volume~35, pp.\  4027--4035, 2021.

\bibitem[Devlin et~al.(2018)Devlin, Chang, Lee, and Toutanova]{devlin2018bert}
Jacob Devlin, Ming-Wei Chang, Kenton Lee, and Kristina Toutanova.
\newblock Bert: Pre-training of deep bidirectional transformers for language
  understanding.
\newblock \emph{arXiv preprint arXiv:1810.04805}, 2018.

\bibitem[Dosovitskiy et~al.(2020)Dosovitskiy, Beyer, Kolesnikov, Weissenborn,
  Zhai, Unterthiner, Dehghani, Minderer, Heigold, Gelly,
  et~al.]{dosovitskiy2020image}
Alexey Dosovitskiy, Lucas Beyer, Alexander Kolesnikov, Dirk Weissenborn,
  Xiaohua Zhai, Thomas Unterthiner, Mostafa Dehghani, Matthias Minderer, Georg
  Heigold, Sylvain Gelly, et~al.
\newblock An image is worth 16x16 words: Transformers for image recognition at
  scale.
\newblock \emph{arXiv preprint arXiv:2010.11929}, 2020.

\bibitem[Goh et~al.(2017)Goh, Adepu, Junejo, and Mathur]{goh2017dataset}
Jonathan Goh, Sridhar Adepu, Khurum~Nazir Junejo, and Aditya Mathur.
\newblock A dataset to support research in the design of secure water treatment
  systems.
\newblock In \emph{International conference on critical information
  infrastructures security}, pp.\  88--99. Springer, 2017.

\bibitem[Goodfellow et~al.(2014)Goodfellow, Pouget-Abadie, Mirza, Xu,
  Warde-Farley, Ozair, Courville, and Bengio]{goodfellow2014generative}
Ian Goodfellow, Jean Pouget-Abadie, Mehdi Mirza, Bing Xu, David Warde-Farley,
  Sherjil Ozair, Aaron Courville, and Yoshua Bengio.
\newblock Generative adversarial nets.
\newblock \emph{Advances in neural information processing systems}, 27, 2014.

\bibitem[Hundman et~al.(2018)Hundman, Constantinou, Laporte, Colwell, and
  Soderstrom]{hundman2018detecting}
Kyle Hundman, Valentino Constantinou, Christopher Laporte, Ian Colwell, and Tom
  Soderstrom.
\newblock Detecting spacecraft anomalies using lstms and nonparametric dynamic
  thresholding.
\newblock In \emph{Proceedings of the 24th ACM SIGKDD international conference
  on knowledge discovery \& data mining}, pp.\  387--395, 2018.

\bibitem[Joshi et~al.(2020)Joshi, Chen, Liu, Weld, Zettlemoyer, and
  Levy]{joshi2020spanbert}
Mandar Joshi, Danqi Chen, Yinhan Liu, Daniel~S Weld, Luke Zettlemoyer, and Omer
  Levy.
\newblock Spanbert: Improving pre-training by representing and predicting
  spans.
\newblock \emph{Transactions of the Association for Computational Linguistics},
  8:\penalty0 64--77, 2020.

\bibitem[Kim et~al.(2022)Kim, Choi, Choi, Lee, and Yoon]{kim2022towards}
Siwon Kim, Kukjin Choi, Hyun-Soo Choi, Byunghan Lee, and Sungroh Yoon.
\newblock Towards a rigorous evaluation of time-series anomaly detection.
\newblock In \emph{Proceedings of the AAAI Conference on Artificial
  Intelligence}, volume~36, pp.\  7194--7201, 2022.

\bibitem[Kingma \& Ba(2014)Kingma and Ba]{kingma2014adam}
Diederik~P Kingma and Jimmy Ba.
\newblock Adam: A method for stochastic optimization.
\newblock \emph{arXiv preprint arXiv:1412.6980}, 2014.

\bibitem[Kriegel et~al.(2009)Kriegel, Kr{\"o}ger, Schubert, and
  Zimek]{kriegel2009loop}
Hans-Peter Kriegel, Peer Kr{\"o}ger, Erich Schubert, and Arthur Zimek.
\newblock Loop: local outlier probabilities.
\newblock In \emph{Proceedings of the 18th ACM conference on Information and
  knowledge management}, pp.\  1649--1652, 2009.

\bibitem[Lai et~al.(2021)Lai, Zha, Xu, Zhao, Wang, and Hu]{lai2021revisiting}
Kwei-Herng Lai, Daochen Zha, Junjie Xu, Yue Zhao, Guanchu Wang, and Xia Hu.
\newblock Revisiting time series outlier detection: Definitions and benchmarks.
\newblock In \emph{Thirty-fifth Conference on Neural Information Processing
  Systems Datasets and Benchmarks Track (Round 1)}, 2021.

\bibitem[Lewis et~al.(2019)Lewis, Liu, Goyal, Ghazvininejad, Mohamed, Levy,
  Stoyanov, and Zettlemoyer]{lewis2019bart}
Mike Lewis, Yinhan Liu, Naman Goyal, Marjan Ghazvininejad, Abdelrahman Mohamed,
  Omer Levy, Ves Stoyanov, and Luke Zettlemoyer.
\newblock Bart: Denoising sequence-to-sequence pre-training for natural
  language generation, translation, and comprehension.
\newblock \emph{arXiv preprint arXiv:1910.13461}, 2019.

\bibitem[Liu et~al.(2008)Liu, Ting, and Zhou]{liu2008isolation}
Fei~Tony Liu, Kai~Ming Ting, and Zhi-Hua Zhou.
\newblock Isolation forest.
\newblock In \emph{2008 eighth ieee international conference on data mining},
  pp.\  413--422. IEEE, 2008.

\bibitem[Liu et~al.(2021)Liu, Lin, Cao, Hu, Wei, Zhang, Lin, and
  Guo]{liu2021swin}
Ze~Liu, Yutong Lin, Yue Cao, Han Hu, Yixuan Wei, Zheng Zhang, Stephen Lin, and
  Baining Guo.
\newblock Swin transformer: Hierarchical vision transformer using shifted
  windows.
\newblock In \emph{Proceedings of the IEEE/CVF International Conference on
  Computer Vision}, pp.\  10012--10022, 2021.

\bibitem[Park et~al.(2018)Park, Hoshi, and Kemp]{park2018multimodal}
Daehyung Park, Yuuna Hoshi, and Charles~C Kemp.
\newblock A multimodal anomaly detector for robot-assisted feeding using an
  lstm-based variational autoencoder.
\newblock \emph{IEEE Robotics and Automation Letters}, 3\penalty0 (3):\penalty0
  1544--1551, 2018.

\bibitem[Raffel et~al.(2019)Raffel, Shazeer, Roberts, Lee, Narang, Matena,
  Zhou, Li, and Liu]{raffel2019exploring}
Colin Raffel, Noam Shazeer, Adam Roberts, Katherine Lee, Sharan Narang, Michael
  Matena, Yanqi Zhou, Wei Li, and Peter~J Liu.
\newblock Exploring the limits of transfer learning with a unified text-to-text
  transformer.
\newblock \emph{arXiv preprint arXiv:1910.10683}, 2019.

\bibitem[Ruff et~al.(2018)Ruff, Vandermeulen, Goernitz, Deecke, Siddiqui,
  Binder, M{\"u}ller, and Kloft]{ruff2018deep}
Lukas Ruff, Robert Vandermeulen, Nico Goernitz, Lucas Deecke, Shoaib~Ahmed
  Siddiqui, Alexander Binder, Emmanuel M{\"u}ller, and Marius Kloft.
\newblock Deep one-class classification.
\newblock In \emph{International conference on machine learning}, pp.\
  4393--4402. PMLR, 2018.

\bibitem[Shen et~al.(2020)Shen, Li, and Kwok]{shen2020timeseries}
Lifeng Shen, Zhuocong Li, and James Kwok.
\newblock Timeseries anomaly detection using temporal hierarchical one-class
  network.
\newblock \emph{Advances in Neural Information Processing Systems},
  33:\penalty0 13016--13026, 2020.

\bibitem[Su et~al.(2019)Su, Zhao, Niu, Liu, Sun, and Pei]{su2019robust}
Ya~Su, Youjian Zhao, Chenhao Niu, Rong Liu, Wei Sun, and Dan Pei.
\newblock Robust anomaly detection for multivariate time series through
  stochastic recurrent neural network.
\newblock In \emph{Proceedings of the 25th ACM SIGKDD international conference
  on knowledge discovery \& data mining}, pp.\  2828--2837, 2019.

\bibitem[Tang et~al.(2002)Tang, Chen, Fu, and Cheung]{tang2002enhancing}
Jian Tang, Zhixiang Chen, Ada Wai-Chee Fu, and David~W Cheung.
\newblock Enhancing effectiveness of outlier detections for low density
  patterns.
\newblock In \emph{Pacific-Asia Conference on Knowledge Discovery and Data
  Mining}, pp.\  535--548. Springer, 2002.

\bibitem[Tax \& Duin(2004)Tax and Duin]{tax2004support}
David~MJ Tax and Robert~PW Duin.
\newblock Support vector data description.
\newblock \emph{Machine learning}, 54\penalty0 (1):\penalty0 45--66, 2004.

\bibitem[Vaswani et~al.(2017)Vaswani, Shazeer, Parmar, Uszkoreit, Jones, Gomez,
  Kaiser, and Polosukhin]{vaswani2017attention}
Ashish Vaswani, Noam Shazeer, Niki Parmar, Jakob Uszkoreit, Llion Jones,
  Aidan~N Gomez, {\L}ukasz Kaiser, and Illia Polosukhin.
\newblock Attention is all you need.
\newblock \emph{Advances in neural information processing systems}, 30, 2017.

\bibitem[Xu et~al.(2018)Xu, Chen, Zhao, Li, Bu, Li, Liu, Zhao, Pei, Feng,
  et~al.]{xu2018unsupervised}
Haowen Xu, Wenxiao Chen, Nengwen Zhao, Zeyan Li, Jiahao Bu, Zhihan Li, Ying
  Liu, Youjian Zhao, Dan Pei, Yang Feng, et~al.
\newblock Unsupervised anomaly detection via variational auto-encoder for
  seasonal kpis in web applications.
\newblock In \emph{Proceedings of the 2018 world wide web conference}, pp.\
  187--196, 2018.

\bibitem[Xu et~al.(2021)Xu, Wu, Wang, and Long]{xu2021anomaly}
Jiehui Xu, Haixu Wu, Jianmin Wang, and Mingsheng Long.
\newblock Anomaly transformer: Time series anomaly detection with association
  discrepancy.
\newblock \emph{arXiv preprint arXiv:2110.02642}, 2021.

\bibitem[Yang et~al.(2019)Yang, Dai, Yang, Carbonell, Salakhutdinov, and
  Le]{yang2019xlnet}
Zhilin Yang, Zihang Dai, Yiming Yang, Jaime Carbonell, Russ~R Salakhutdinov,
  and Quoc~V Le.
\newblock Xlnet: Generalized autoregressive pretraining for language
  understanding.
\newblock \emph{Advances in neural information processing systems}, 32, 2019.

\bibitem[Zhang et~al.(2019)Zhang, Song, Chen, Feng, Lumezanu, Cheng, Ni, Zong,
  Chen, and Chawla]{zhang2019deep}
Chuxu Zhang, Dongjin Song, Yuncong Chen, Xinyang Feng, Cristian Lumezanu, Wei
  Cheng, Jingchao Ni, Bo~Zong, Haifeng Chen, and Nitesh~V Chawla.
\newblock A deep neural network for unsupervised anomaly detection and
  diagnosis in multivariate time series data.
\newblock In \emph{Proceedings of the AAAI conference on artificial
  intelligence}, volume~33, pp.\  1409--1416, 2019.

\bibitem[Zong et~al.(2018)Zong, Song, Min, Cheng, Lumezanu, Cho, and
  Chen]{zong2018deep}
Bo~Zong, Qi~Song, Martin~Renqiang Min, Wei Cheng, Cristian Lumezanu, Daeki Cho,
  and Haifeng Chen.
\newblock Deep autoencoding gaussian mixture model for unsupervised anomaly
  detection.
\newblock In \emph{International conference on learning representations}, 2018.

\end{thebibliography}
\bibliographystyle{iclr2023_conference}


\appendix
\section{Experimental Details}

\subsection{Detailed Information of Datasets}
\label{sec:appendix_datasets}

We create the simple sine wave dataset according to the synthesizing criteria introduced by \citet{lai2021revisiting}.
Based on a noised sine wave, we split the wave into a long length of a normal sequence for training and five slices of abnormal sequences for testing.
Each abnormal slice contains one of the typical types of outliers; global, contextual, shapelet, seasonal, and trend outliers.

\begin{table}[h]
\renewcommand{\arraystretch}{1.1}
\caption{A summary of five real-world benchmark datasets. * indicates the average length of all sub-datasets in the case of SMD.}
\label{tab:data_info}
\vspace*{-2mm}
\begin{center}
\begin{tabular}{c|cccc}
Dataset & Train length & Test length & Anomaly \% in test & Dimension \\
\hline
SWaT \citeyearpar{goh2017dataset} & 495,000 & 449,919 & 12.13\% & 51 \\
WADI \citeyearpar{ahmed2017wadi} & 784,537 & 172,801 & 5.77\% & 123 \\
MSL \citeyearpar{hundman2018detecting} & 58,317 & 73,729 & 10.53\% & 55 \\
SMAP \citeyearpar{hundman2018detecting} & 153,183 & 427,617 & 12.79\% & 25 \\
SMD \citeyearpar{su2019robust} & 25,300$^*$ & 25,301$^*$ & 4.16\% & 38
\end{tabular}
\end{center}
\end{table}

The five real-world benchmark datasets are summarized in Table \ref{tab:data_info} and described below.

{\bf Secure Water Treatment (SWaT)} \citep{goh2017dataset}.
SWaT is collected from a water treatment testbed for seven days under normal conditions and four days with physical attacks.
The data is composed of 51 sensor values.
In practice, we ignore the eleventh column in the entire data because unseen patterns labeled as normal in the testing part arise occasionally out of the range of values in the training part.
(Our method detects these kinds of normal patterns as anomalies.)

{\bf Water Distribution Testbed (WADI)} \citep{ahmed2017wadi}.
WADI is a dataset collected from a water distribution testbed with 123 sensors.
It contains a normal sequence of 14 days and an abnormal sequence of two days with attack scenarios.
For the same reason as SWaT, we ignore the 102nd column in the entire data.

{\bf Mars Science Laboratory (MSL)} \citep{hundman2018detecting}.
MSL is telemetry data collected by the NASA spacecraft.
It consists of datasets from 55 telemetry channels, and each test set is labeled from Incident Surprise Anomaly reports.
Following \citet{su2019robust}, we concatenate all datasets into a pair of training and testing sets and yield a single result.
We also prevent windows containing data from multiple channels for score prediction.

{\bf Soil Moisture Active Passive (SMAP)} \citep{hundman2018detecting}.
SMAP is also collected by the NASA spacecraft and has a similar characteristic to MSL.
All sub-datasets are concatenated and produce a single result as in MSL.

{\bf Server Machine Dataset (SMD)} \citep{su2019robust}.
SMD is a collection of sub-datasets from 28 different machines provided by a large Internet company.
Each sub-dataset is equally divided into two parts, the first half for training and the second half for testing.
The training and evaluation procedures are carried out on 28 sub-datasets separately, and the averaged results are used for comparisons.

\subsection{Training and Evaluation Settings}
\label{sec:appendix_settings}
We use one linear layer as the embedding layer and 2-layer MLPs as the prediction block.
The Transformer body has six layers of the embedding dimension of 512 and eight attention heads.
The prediction block contains one hidden layer of 2,048 neurons with GELU activation in between.
The sequence length of embedded features is 512 but window sizes (and patch sizes) of input data vary with datasets.
For the preliminary experiment in Section \ref{sec:motivation}, we simplify our model by halving the embedding dimension and the number of Transformer layers and shortening the window size and patch size to 100 and 1, respectively.

In the training stage, we train the model with the mini-batch size of 16 for the maximum training steps of 150K.
Input windows are selected randomly from the training set at every step.
External sequences for the soft replacement are also selected randomly from the same training set.
When degrading a window, an average of 30\% of columns are degraded but the remaining columns are left.
A synthetic outlier type is selected as the soft replacement in 50\% probability, uniform replacement in 15\% probability, peak noise in 15\% probability, and length adjustment in 10\% probability.
However, the length adjustment may not be used for several datasets because it may reduce the model capability depending on the datasets.
The other settings depending on the datasets are presented in Table \ref{tab:settings_dataset}.
We employ the AdamW optimizer \citep{kingma2014adam} with a learning rate of $1 \times 10^{-4}$, and use a learning rate warmup for 10\% of training steps and a cosine learning rate decay.
To prevent exploding gradients, gradient clipping is applied at a norm of 1.0.
For the score prediction of the test set, we slide input windows with the sliding step of 16.

\begin{table}[h]
\renewcommand{\arraystretch}{1.1}
\caption{Different training settings for five benchmark datasets.}
\label{tab:settings_dataset}
\vspace*{-2mm}
\begin{center}
\begin{tabular}{c|ccccc}
\diagbox[width=11.5em]{Setting}{Dataset} & SWaT & WADI & MSL & SMAP & SMD \\
\hline
Patch size & 14 & 8 & 2 & 4 & 4 \\
Window size & 7,168 & 4,096 & 1,024 & 2,048 & 2,048 \\
Max length \% of outlier & 20\% & 15\% & 20\% & 15\% & 20\% \\
Use of length adjustment & \circbox0{cblue} & \textcolor{lightred}{$\times$} & \textcolor{lightred}{$\times$} & \textcolor{lightred}{$\times$} & \circbox0{cblue}
\end{tabular}
\end{center}
\end{table}

\subsection{Detailed Results of Section \ref{sec:motivation}}
We examine relationships between the typical types of outliers and our proposed synthetic outliers in Section \ref{sec:motivation} through $\mathrm{F1}$ and AUROC.
The detailed results of $\mathrm{F1}$ and AUROC are presented in Table \ref{tab:Motivation_results_F1} and Table \ref{tab:Motivation_results_AUROC}, respectively.

\begin{table}[h]
\renewcommand{\arraystretch}{1.1}
\caption{AUROC results of the preliminary experiment in Section \ref{sec:motivation}.}
\label{tab:Motivation_results_AUROC}
\vspace*{-2mm}
\begin{center}
\begin{tabular}{l|ccccc}
\multirow{2}{*}{\diagbox{Proposed}{Typical}}
& \multicolumn{2}{c}{\bf Point} & \multicolumn{3}{c}{\bf Pattern} \\
                    &       Global       &    Contextual    &     Shapelet     &     Seasonal      &     Trend          \\ \hline
Soft replacement    &  {\bf 1.000}       &  {\bf 1.000}     &  {\bf 1.000}     &  {\bf 0.997 }     &  {\bf 1.000}       \\
Uniform replacement &       0.419        &  {\bf 1.000}     &  {\bf 1.000}     &       0.927       &      0.899         \\
Length adjustment   &       0.990        &      0.965       &        0.894     &  {\bf 0.998 }     &      0.551         \\
Peak noise          &  {\bf 1.000}       &  {\bf 1.000}     &       0.863      &       0.847       &      0.800
\end{tabular}
\end{center}
\end{table}

\begin{table}[h]
\renewcommand{\arraystretch}{1.1}
\caption{$\mathrm{F1}$ results of the preliminary experiment in Section \ref{sec:motivation}.}
\label{tab:Motivation_results_F1}
\vspace*{-2mm}
\begin{center}
\begin{tabular}{l|ccccc}
\multirow{2}{*}{\diagbox{Proposed}{Typical}}
& \multicolumn{2}{c}{\bf Point} & \multicolumn{3}{c}{\bf Pattern} \\
                    &       Global       &    Contextual    &     Shapelet     &     Seasonal      &     Trend          \\ \hline
Soft replacement    &  {\bf 1.000}       &  {\bf 1.000}     &  {\bf 1.000}     &{\bf 0.952 }       &  {\bf 1.000}       \\
Uniform replacement &   0.000            &  {\bf 1.000}     &  {\bf 1.000}     &   0.825           &   0.889             \\
Length adjustment   &   0.667            &   0.400          &   0.556          &{\bf 0.947}         &   0.625              \\
Peak noise          &  {\bf 1.000}       &  {\bf 1.000}     &   0.588          &   0.824           &   0.824          
\end{tabular}
\end{center}
\end{table}


\section{Visual Results}

\begin{figure}[h]
\begin{center}
\includegraphics[width=\textwidth]{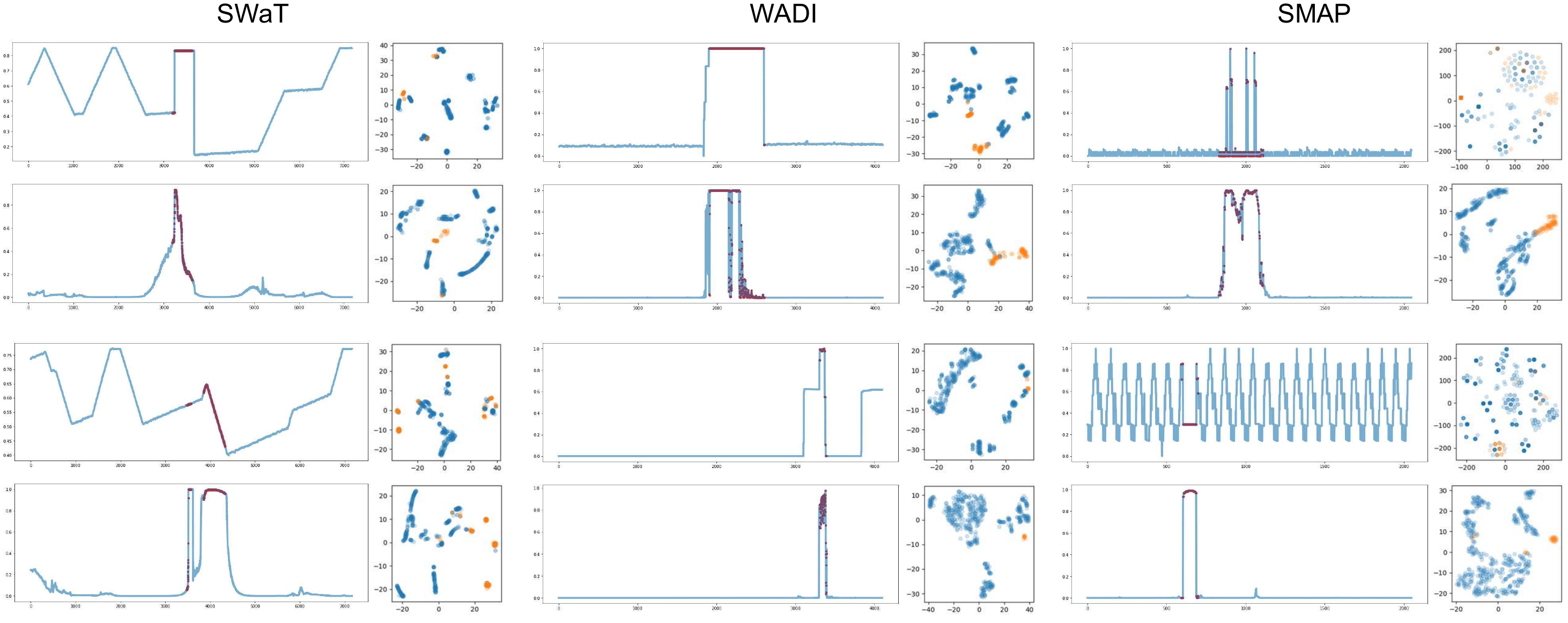}
\includegraphics[width=\textwidth]{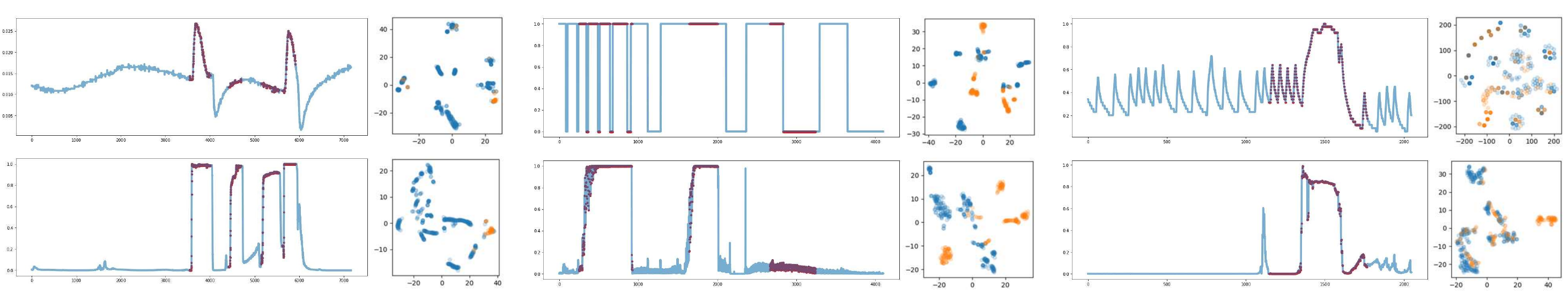}
\end{center}
\setlength{\abovecaptionskip}{5pt plus 3pt minus 2pt}
\caption{Visualization of anomaly score predictions on abnormal sequences in SWaT, WADI, and SMAP datasets. In each subfigure, graphs of an abnormal component where an outlier occurs ({\it top left}) and the corresponding anomaly scores ({\it bottom left}) are presented, and the data points ({\it top right}) and the last latent features ({\it bottom right}) are visualized in 2D planes. Blue lines/dots represent normal values and red/orange dots represent abnormal values.}
\label{fig:visualization}
\end{figure}

We visualize the prediction results of AnomalyBERT and present the 2D projections of the original data and latent features in the model.
Similar to Figure \ref{fig:introduction_examples}, we select three abnormal windows in each of SWaT, WADI, and SMAP test sets and fetch the corresponding anomaly scores after the score prediction process.
We then project the original data points and the last latent features from the abnormal windows into 2D planes using t-SNE.
As shown in Figure \ref{fig:visualization}, our method distinguishes anomalies successfully and separates abnormal data points from normal points well.

\end{document}